\documentclass[conference]{IEEEtran}
\IEEEoverridecommandlockouts

\usepackage{cite}
\usepackage{amsmath,amssymb,amsfonts}
\usepackage{algorithmic}
\usepackage{graphicx}
\usepackage{textcomp}
\usepackage{xcolor}
\usepackage{graphicx}
\usepackage{hyperref}       
\usepackage{url}            
\usepackage{booktabs}       
\usepackage{amsfonts}       
\usepackage{nicefrac}       
\usepackage{microtype}      
\usepackage{lipsum}
\usepackage{fancyhdr}       
\usepackage{float}
\usepackage{subcaption}
\usepackage{caption}
\usepackage{adjustbox}
\usepackage{makecell}
\usepackage{tikz}
\usepackage{pgfplots}
\usepackage{amsmath}
\usepackage{svg}
\usepackage{setspace}
\usepackage{amsmath}
\usepackage{graphicx}
\usepackage{hyperref}
\usepackage{multicol}
\usepackage{enumitem}

\def\BibTeX{{\rm B\kern-.05em{\sc i\kern-.025em b}\kern-.08em
    T\kern-.1667em\lower.7ex\hbox{E}\kern-.125emX}}
\begin{document}

\title{Mitigating Overfitting in Medical Imaging: Self-Supervised Pretraining vs. ImageNet Transfer Learning for Dermatological Diagnosis}

\author{
\IEEEauthorblockN{Iván Matas}
\IEEEauthorblockA{\textit{Dpt. of Signal Theory and Communications} \\ 
\textit{University of Seville} \\ 
Seville, Spain \\ 
imatas@us.es}
\and
\IEEEauthorblockN{Carmen Serrano}
\IEEEauthorblockA{\textit{Dpt. of Signal Theory and Communications} \\ 
\textit{University of Seville} \\ 
Seville, Spain \\ 
cserrano@us.es}
\and
\IEEEauthorblockN{Miguel Nogales}
\IEEEauthorblockA{\textit{Dpt. of Signal Theory and Communications} \\ 
\textit{University of Seville} \\ 
Seville, Spain \\ 
mnogales@us.es}
\and
\IEEEauthorblockN{David Moreno}
\IEEEauthorblockA{\textit{Department of Dermatology} \\ 
\textit{Hospital Universitario Virgen Macarena} \\ 
Seville, Spain \\ 
david.moreno.ramirez.sspa@juntadeandalucia.es}
\and
\IEEEauthorblockN{Lara Ferrándiz}
\IEEEauthorblockA{\textit{Department of Dermatology} \\ 
\textit{Hospital Universitario Virgen Macarena} \\ 
Seville, Spain \\ 
lara.ferrandiz.sspa@juntadeandalucia.es}
\and
\IEEEauthorblockN{Teresa Ojeda}
\IEEEauthorblockA{\textit{Department of Dermatology} \\ 
\textit{Hospital Quirón Salud Sevilla} \\ 
Seville, Spain \\ 
tojedavila@gmail.com}
\and
\IEEEauthorblockN{Begoña Acha}
\IEEEauthorblockA{\textit{Dpt. of Signal Theory and Communications} \\ 
\textit{University of Seville} \\ 
Seville, Spain \\ 
bacha@us.es}
}

\maketitle

\begin{abstract}
Deep learning has transformed computer vision but relies heavily on large labeled datasets and computational resources. Transfer learning, particularly fine-tuning pretrained models, offers a practical alternative; however, models pretrained on natural image datasets such as ImageNet may fail to capture domain-specific characteristics in medical imaging. 
This study introduces an unsupervised learning framework that extracts high-value dermatological features instead of relying solely on ImageNet-based pretraining. We employ a Variational Autoencoder (VAE) trained from scratch on a proprietary dermatological dataset, allowing the model to learn a structured and clinically relevant latent space. This self-supervised feature extractor is then compared to an ImageNet-pretrained backbone under identical classification conditions, highlighting the trade-offs between general-purpose and domain-specific pretraining.
Our results reveal distinct learning patterns. The self-supervised model achieves a final validation loss of 0.110 (-33.33\%), while the ImageNet-pretrained model stagnates at 0.100 (-16.67\%), indicating overfitting. Accuracy trends confirm this: the self-supervised model improves from 45\% to 65\% (+44.44\%) with a near-zero overfitting gap, whereas the ImageNet-pretrained model reaches 87\% (+50.00\%) but plateaus at 75\% (+19.05\%), with its overfitting gap increasing to +0.060. 
These findings suggest that while ImageNet pretraining accelerates convergence, it also amplifies overfitting on non-clinically relevant features. In contrast, self-supervised learning achieves steady improvements, stronger generalization, and superior adaptability, underscoring the importance of domain-specific feature extraction in medical imaging.

\end{abstract}

\begin{IEEEkeywords}
Transfer learning, deep learning, self-supervised learning, ImageNet, domain adaptation, medical imaging, variational autoencoder, dermatological classification.
\end{IEEEkeywords}

\section{Introduction}

Deep learning has emerged as a transformative force in computer vision, facilitating state-of-the-art performance in tasks such as image classification, object detection, and segmentation. However, the inherent computational expense and extensive labeled datasets required to train deep neural networks from scratch render this approach infeasible in many practical applications. To address these challenges, fine-tuning has become a pivotal methodology within transfer learning, enabling the adaptation of pretrained models to specific domains while mitigating the burden of training large-scale models from the ground up. Fine-tuning leverages the knowledge from a large source dataset—typically a corpus of natural images. This approach accelerates convergence, enhances generalization, and improves performance with fewer labeled samples.

A widely adopted paradigm in fine-tuning involves initializing models with weights pretrained on extensive datasets such as ImageNet, which encompasses a diverse collection of natural images. These pretrained architectures capture hierarchical feature representations, which can be repurposed for domain-specific applications with limited training data. Nevertheless, the effectiveness of transfer learning is fundamentally contingent upon the degree of similarity between the source and target domains. When the target dataset exhibits significant divergence from ImageNet, conventional fine-tuning strategies may lead to suboptimal generalization and increased susceptibility to overfitting. This limitation underscores the necessity of exploring the comparative advantages of using ImageNet pretraining versus domain-specific pretraining, particularly in specialized fields such as medical imaging and fine-grained visual categorization.

Recent studies have examined the limitations and benefits of ImageNet pretraining in medical imaging. Juodelyte et al. and Raptis et al. \cite{juodelyte2025, raptis2024} found that while ImageNet-pretrained models achieved competitive performance, they were prone to overfitting to spurious correlations, reducing their robustness to out-of-distribution samples. Their analyses, spanning X-ray, CT, cytological, and MRI scans, suggest that models pretrained on domain-specific datasets often outperform ImageNet-based counterparts, particularly when feature distributions diverge significantly from natural images. Similarly, Zhang et al. \cite{zhang2023} demonstrated that ImageNet pretraining accelerates convergence and improves early-stage accuracy, yielding a 3–4\% classification accuracy gain over training from scratch. These findings highlight both the efficiency advantages of transfer learning and its fundamental limitations when applied to specialized medical domains.

Kornblith et al. \cite{kornblith2024} evaluated ImageNet-pretrained models across twelve datasets, finding a strong correlation ($r = 0.96$) between ImageNet accuracy and transfer learning performance. While superior ImageNet models generally improved fine-tuning outcomes, their study also revealed that for small, fine-grained classification tasks, ImageNet pretraining offered minimal benefits, indicating limited generalizability to highly specialized domains.

These findings highlight the critical role of pretraining strategies in model robustness and generalization. While ImageNet pretraining remains a widely used baseline, models trained on domain-specific datasets demonstrate greater resilience to dataset biases and domain shifts. This underscores the trade-off between broad cross-domain generalization and specialization for specific tasks, an ongoing challenge in transfer learning research.

A key limitation of ImageNet-pretrained models is their tendency to overfit more quickly than models pretrained on datasets closely aligned with the target domain. Although ImageNet pretraining provides a strong initialization, its feature representations may be suboptimal for highly specialized tasks with distinct data distributions. Addressing this issue requires tailored fine-tuning strategies that enhance feature adaptation while retaining the advantages of pretraining.

In this study, we systematically analyze the overfitting behavior of ImageNet-pretrained models compared to those pretrained on domain-specific datasets. By evaluating performance across architectures, datasets, and training conditions, we identify the inherent limitations of conventional transfer learning and propose strategies to improve generalization in specialized applications. Our findings contribute to the ongoing discourse on transfer learning, offering insights into optimizing pretraining approaches for enhanced model robustness and performance.

\subsection{Rationale and Advantages}
The decision to train a model from scratch using unsupervised learning offers several critical advantages in medical imaging applications. Domain-specific feature extraction enables the identification of intrinsic patterns within dermatological data, capturing nuanced features that might elude human annotation. This is particularly valuable for distinguishing between complex conditions such as Basal Cell Carcinoma (BCC), melanoma, actinic keratosis, squamous cell carcinoma (SCC),and so on.
By initializing with random weights rather than pretrained parameters from ImageNet (which predominantly contains non-medical images), the model constructs hierarchical feature representations specific to dermatological imaging, avoiding the introduction of biases from unrelated visual domains. The model develops representations uniquely tailored to the hospital's imaging techniques and patient demographics, enhancing real-world clinical applicability.
Unlike supervised approaches that may overfit to predefined labels, our unsupervised methodology focuses on learning the inherent structure of skin lesion data, reducing the risk of capturing superficial or clinically irrelevant patterns.

\subsection{Clinical Implications}
This approach addresses a fundamental challenge in medical image classification: the extraction of weak or non-informative features that can compromise diagnostic accuracy. By focusing on the intrinsic characteristics of dermatological lesions, the model distinguishes critical visual markers essential for accurate diagnosis, improving both classification performance and clinical interpretability.
The enhanced feature extraction capability facilitates better generalization across diverse dermatological conditions encountered in clinical practice, potentially improving diagnostic support in ambiguous cases where visual distinction between pathologies requires sophisticated pattern recognition.

\section{Material} 

\subsection{Database}\label{sec:BBDD}

The dataset used in this study originates from ``Hospital Universitario Virgen Macarena'', a public hospital from the Andalusian Healthcare System, located in Seville, Spain. It is an extensive collection of dermatological cases compiled from 60 primary healthcare centers, where clinical data and images have been gathered over multiple years from numerous patients presenting various skin lesions. These images are sent to the main hospital via teledermatology.

The dataset contained a total of 200,000 images, with approximately 40\% being clinical images (capturing a broader anatomical context) and 60\% being dermatoscopic images (focused on skin lesions). Given the study's objective, only the dermatoscopic images were retained for further processing. The dataset comprises 15 distinct skin lesion classes, which include the most common dermatological conditions related to skin cancer and other skin pathologies; details can be found in Table \ref{tab:grouping_dermatological_classes_priority}.



To enhance the clinical relevance of the classification model, the hospital established a priority-based grouping of dermatological lesion types. This categorization, presented in the following table, stratifies lesions according to their urgency for medical intervention. By structuring the classes into three priority levels, the classification framework aligns with the practical demands of dermatological diagnosis, ensuring that high-risk conditions receive prompt attention while maintaining a systematic approach to less urgent cases.

\begin{table}
    \centering
    \resizebox{0.98\columnwidth}{!}{ 
    \begin{tabular}{m{3cm} m{6cm}} 
        \toprule
        \textbf{Priority} & \hspace{2cm}\textbf{Classes} \\
        \midrule
        \textbf{Priority 1} & 
        \begin{itemize}
            \setlength\itemsep{0.2cm} 
            \item Melanoma
            \item Squamous Cell Carcinoma
            \item Basal Cell Carcinoma
            \item Superficial Basal Cell Carcinoma
        \end{itemize} \\
        \midrule
        \textbf{Priority 2} & 
        \begin{itemize}
            \setlength\itemsep{0.2cm}
            \item Actinic Keratosis
            \item Common Acquired Melanocytic \mbox{Nevus}
            \item Atypical Melanocytic Nevus
            \item Acral Melanocytic Nevus
            \item Spitz Reed Nevus
            \item Irritated Melanocytic Nevus
        \end{itemize} \\
        \midrule
        \textbf{Priority 3} & 
        \begin{itemize}
            \setlength\itemsep{0.2cm}
            \item Acquired Angioma
            \item Dermatofibroma
            \item Other Skin Lesions
        \end{itemize} \\
        \bottomrule
    \end{tabular}
    } 
    \caption{Prioritization of dermatological categories for a \\ triage tool at ``Hospital Universitario Virgen Macarena''. Priority 1 is the most severe class while priority 3 is the lowest.}
    \label{tab:grouping_dermatological_classes_priority}
\end{table}



To address the dataset's class imbalance and ensure robust model generalization, a preprocessing pipeline was applied. Key steps included removing low-quality images, eliminating redundant patient samples to reduce bias, and retaining only high-quality dermatological images for diagnostic consistency.

To further mitigate domain shift issues and enhance model robustness, we integrated an additional dataset: the well-established ISIC Challenge dataset, widely recognized in dermatological research. The lesion classes from both sources were mapped to the priority classification scheme provided by the hospital. 

This integration was crucial, as early training stages revealed that models trained on a single dataset could differentiate image sources rather than focusing on the anatomical and pathological features required for lesion classification. By merging datasets and aligning lesion priorities, we ensured a more generalizable and clinically reliable model.

\section{
    Methodology and Technical Implementation
} 

\begin{figure}[h!]
    \centering
    \includegraphics[width = \columnwidth]{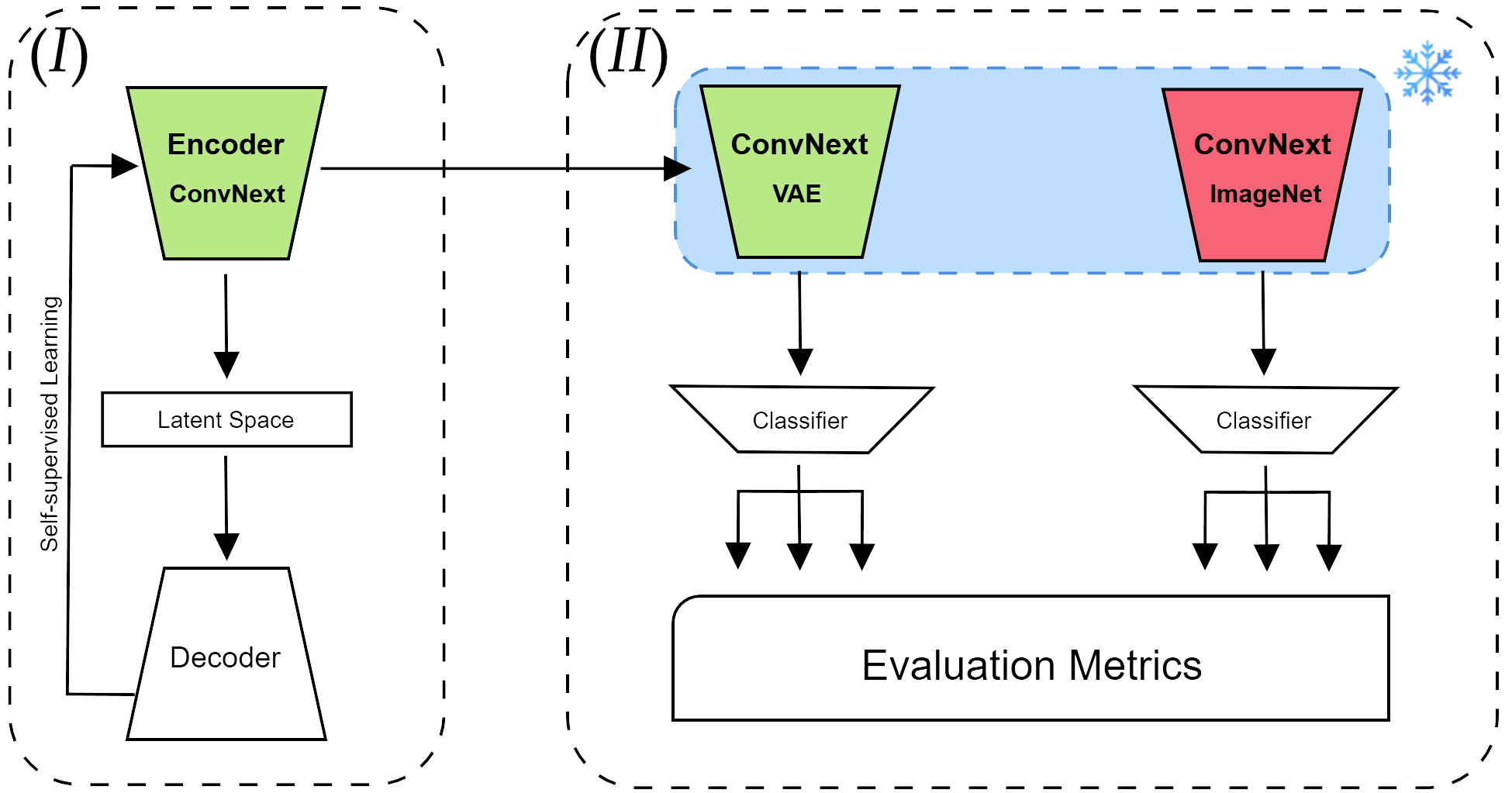}
    \caption{Overview of the proposed self-supervised learning framework. (I) The Variational AutoEncoder is trained in an unsupervised manner, using a randomly initialized ConvNext-Tiny encoder to extract robust feature representations. (II) The trained encoder is then frozen and used as a feature extractor for a classification task. A parallel comparison is conducted using a ConvNext-Tiny encoder pretrained on ImageNet. Both feature extractors feed into identical classifier architectures, and their performance is evaluated using the same metrics.}
    \label{fig:Scheme}
\end{figure}

We developed an unsupervised learning framework based on a Variational AutoEncoder (VAE) to extract robust, domain-specific feature representations from dermatological images. Our approach deviates from conventional reliance on ImageNet pretraining by employing a ConvNext-Tiny encoder with randomly initialized weights, ensuring that the learned features are tailored exclusively to the dermatological domain. The model was trained for 300 epochs on a proprietary dataset provided by ``Hospital Universitario Virgen Macarena'' in Seville. The overall methodology, illustrated in Fig. \ref{fig:Scheme}, comprises two key stages: (I) self-supervised training of the VAE, where the model learns to encode and reconstruct dermatological images, and (II) a subsequent classification task that evaluates the effectiveness of the learned feature representations. Each stage is described in detail below.

\begin{enumerate}[label=(\Roman*)]  
    \item \textbf{Self-Supervised Training} \\
    For the development of a self-supervised model, a VAE architecture \cite{VAE} was employed, utilizing ConvNext Tiny \cite{liu2022convnet} as the encoder, with its weights initialized randomly. The latent space was designed with 256 neurons for both $\mu$ and $\log(\sigma^2)$, ensuring a structured and high-capacity representation. The decoder was manually constructed as a symmetric counterpart to the encoder, preserving the architectural principles of ConvNext, but adapted for upsampling instead of downsampling. 

    To optimize the model, the Evidence Lower Bound (ELBO) function was employed, incorporating a warm-up strategy. This method progressively increased the contribution of the Kullback–Leibler divergence term over the course of training, ensuring that while reconstruction remained a fundamental objective, the latent space regularization gained prominence as training advanced. This progressive balancing fosters a more robust representation, allowing the model to encode richer and more informative features from the dataset.

    The training was conducted over a total of 300 epochs using the specified dataset, with an initial learning rate set to $10^{-8}$, ensuring controlled and stable convergence.
    
    \item \textbf{Classification task} \\
    In the second stage, two different methodologies were employed for the classification task and were compared (See Fig.  \ref{fig:Scheme}): 

    \begin{enumerate}
        \item  The encoder, trained in an unsupervised manner with a dermatological database. This feature extractor is specifically tailored to the domain of dermatological imaging, having been trained in an unsupervised manner during the previous stage. This encoder was then frozen and used as a feature extractor for a classification task. 
        \item A parallel ConvNext-Tiny encoder pretrained on ImageNet.
    \end{enumerate}

    Both feature extractors feed into identical classifier architectures, and their performance is evaluated using the same metrics. Both feature extractors, the newly trained one by self-supervised method and the ImageNet-pretrained counterpart, were frozen, preserving all learned weights, and subsequently coupled with a symmetric classifier. This setup ensures that both feature extractors employ an identical classification architecture for the classification task, allowing for a direct comparison between domain-specific self-supervised pretraining and conventional ImageNet-based pretraining, which is the central objective of this study.
    
    The classifier consists of two fully connected layers: the first mapping from 768 to 256 neurons, and the second from 256 to 3 neurons. Rectified Linear Unit (ReLU) activations were applied between layers, alongside a 50\% dropout rate to enhance generalization and mitigate overfitting. Both models were evaluated in parallel under the same set of hyperparameters, ensuring a controlled and comparative assessment. The selected hyperparameters included a total of 30 training epochs, a learning rate of $10^{-5}$, and the AdamW Schedule-Free optimizer \cite{AdamWScheduleFree}. Additionally, the Focal Loss \cite{FocalLoss} function was employed to address the significant class imbalance present in the dataset, as will be discussed in subsequent sections.
\end{enumerate}

\section{Results}

\begin{figure}
    \centering
    \includegraphics[width = \columnwidth]{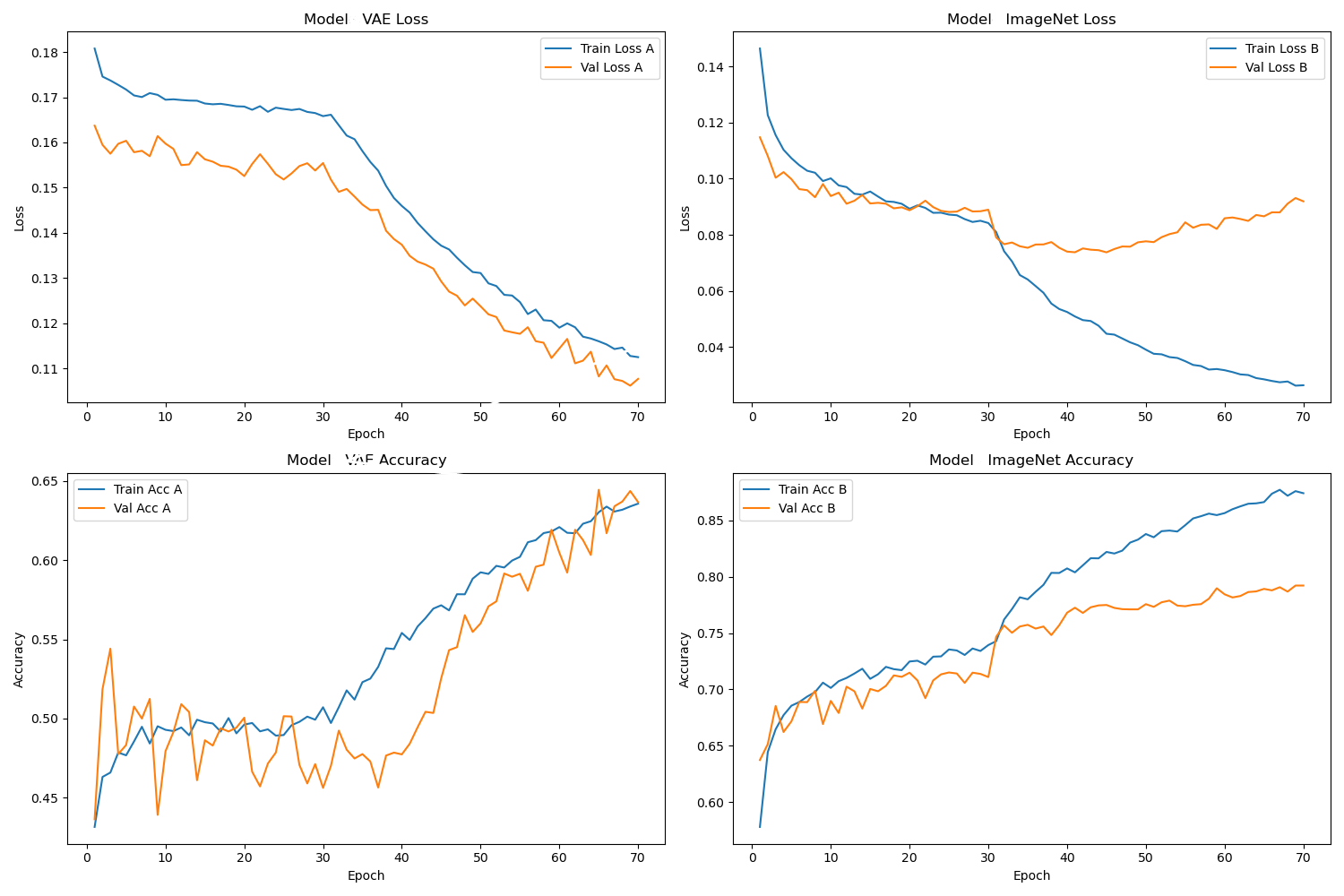}
    \caption{Training dynamics of the self-supervised model (Model A) and the ImageNet-pretrained model (Model B). The left side displays the loss and accuracy evolution of Model A, while the right side shows the corresponding trends for Model B. The top panels illustrate the loss progression, while the bottom panels depict the accuracy evolution.}
    \label{fig:evolution}
\end{figure}

Figure \ref{fig:evolution} illustrates the training dynamics of both models. On the left side, the top panel presents the loss evolution of the backbone trained in an unsupervised manner, while the bottom panel displays its corresponding accuracy progression. On the right side, the top panel shows the loss evolution of the ImageNet-pretrained backbone, and the bottom panel depicts its accuracy trend. This layout provides a direct comparison of both training approaches, highlighting differences in convergence behavior and generalization performance.

During the training process, it was observed that the model utilizing the ImageNet pretrained backbone converged significantly faster and achieved superior performance metrics in a shorter period compared to the model initialized with self-supervised pretrained weights. Additionally, at epoch 30, when the final stage of ConvNext Tiny was fine-tuned in both models, a distinct pattern emerged: the ImageNet pretrained model, despite showing continued improvement in performance metrics, exhibited clear signs of overfitting. In contrast, the model trained with self-supervised weights (VAE) maintained a consistent improvement trajectory without any indication of overfitting.

In order to provide a quantitative perspective on these learning dynamics, we have computed key numerical metrics presented in Table \ref{tab:numerical_results}. The self-supervised model (Model A) demonstrated a steady and progressive learning trend, achieving a final training loss of 0.115 and a validation loss of 0.110, corresponding to a reduction of 36.11\% and 33.33\%, respectively, over 70 epochs. In contrast, the ImageNet-pretrained model (Model B) exhibited a much sharper initial decline in loss, decreasing from 0.145 to 0.040 (-72.41\%). However, its validation loss stagnated at 0.100 (-16.67\%), indicating a lower degree of generalization.

Regarding accuracy, Model A improved from 45\% to 65\% in training accuracy (+44.44\%), while its validation accuracy followed a nearly identical trajectory, confirming a well-regularized learning process. Conversely, Model B reached a higher training accuracy of 87\% (+50.00\%) but plateaued at 75\% validation accuracy (+19.05\%), suggesting an overfitting trend. This is further supported by the overfitting gap, which increased from -0.025 at epoch 0 to +0.060 at epoch 70 for Model B. In contrast, Model A maintained a near-zero difference (-0.005 at epoch 70), reinforcing its superior generalization capabilities.

The learning dynamics further illustrate the differences in optimization behavior. Model B exhibited a steep decline in loss during the first 30 epochs (slope = -0.0047), yet its learning slowed significantly after epoch 30 (slope = 0.0008), indicating early saturation. In contrast, Model A followed a more consistent loss reduction (slope = -0.0003 for epochs 0-29, -0.0014 for fine-tuning epochs), suggesting a more gradual and sustained convergence. A similar pattern was observed in accuracy trends, where Model A maintained a higher accuracy growth rate after epoch 30 (slope = 0.0038) compared to Model B (slope = 0.0030), further highlighting its continued learning potential.

These findings suggest that the self-supervised model achieves a more stable and generalizable representation, avoiding the rapid saturation and overfitting tendencies observed in the ImageNet-pretrained model. While Model B benefits from a strong initial feature extraction capability, its reliance on pre-existing representations limits its ability to adapt to the domain-specific dataset, ultimately leading to diminished validation performance. In contrast, Model A demonstrates superior long-term learning and regularization properties, making it a more robust alternative for real-world applications.

\begin{table*}[t]
    \centering
    \caption{Summary of key numerical results comparing Model A (Self-Supervised) and Model B (ImageNet-Pretrained).}
    \label{tab:numerical_results}
    \begin{tabular}{lcccccc}
        \hline
        \textbf{Metric} & \textbf{Final Value} & \textbf{Overall Change} & \textbf{\% Change} & \textbf{Slope (0-29)} & \textbf{Slope (30-70)} & \textbf{Overfitting at 70} \\
        \hline
        Train Loss A  & 0.115  & -0.065  & -36.11\% & -0.0003 & -0.0014 & -0.005 \\
        Val Loss A    & 0.110  & -0.055  & -33.33\% & -0.0003 & -0.0011 & -0.005 \\
        Train Acc A   & 0.650  & +0.200  & +44.44\% &  0.0017 &  0.0038 & 0.000 \\
        Val Acc A     & 0.650  & +0.200  & +44.44\% &  0.0000 &  0.0050 & 0.000 \\
        \hline
        Train Loss B  & 0.040  & -0.105  & -72.41\% & -0.0047 &  0.0008 & 0.060 \\
        Val Loss B    & 0.100  & -0.020  & -16.67\% & -0.0038 &  0.0023 & 0.060 \\
        Train Acc B   & 0.870  & +0.290  & +50.00\% &  0.0059 &  0.0030 & 0.120 \\
        Val Acc B     & 0.750  & +0.120  & +19.05\% &  0.0024 &  0.0013 & 0.120 \\
        \hline
    \end{tabular}
\end{table*}

\section{Conclusion and Discussion}
Upon completing the training and analyzing the resulting performance metrics, it can be concluded that the ImageNet-based model achieves superior classification performance. However, a more detailed examination of the learning curves suggests that these seemingly better results may be attributed to overfitting on non-clinically relevant features, which may not be meaningful within the medical domain. In contrast, the VAE-based model retains the potential for further enhancement, offering room for improvement through additional training stages and classifier refinements. This suggests that, with appropriate modifications, the self-supervised approach could potentially match or surpass the performance of the ImageNet model while maintaining better generalization to clinically significant features.

The experimental findings presented in this study establish a direct connection between the initial hypotheses and the observed empirical results. At the outset, it was posited that leveraging models pretrained on ImageNet would accelerate convergence and enhance early-stage accuracy, owing to the extraction of general feature representations from a vast corpus of natural images. The experimental data substantiates this claim, as the ImageNet-based model exhibited markedly faster convergence and superior performance during the initial training epochs.

However, this early advantage was counterbalanced by a pronounced tendency toward overfitting. The theoretical framework suggested that transferring learning from a general domain to a highly specialized field such as dermatological imaging could lead the model to excessively adapt to accidental correlations inherent in the pretraining data. This phenomenon was clearly reflected in the learning curves, highlighting the intrinsic limitation of employing a generic pretraining approach when clinically relevant feature extraction is paramount.

Conversely, the self-supervised approach developed through the VAE architecture—where the model was trained from randomly initialized weights tailored to the morphological characteristics of dermatological images—demonstrated a steady improvement without signs of overfitting. This behavior reinforces the premise that domain-specific pretraining is more effective in capturing the subtle and robust features necessary for clinical applications. Although the initial quantitative performance of the self-supervised model was lower, its sustained progression throughout the training process suggests significant potential to eventually match or even surpass the performance of the ImageNet-based model while ensuring that the learned features remain clinically pertinent.

In summary, the comparative analysis of both approaches reveals a delicate balance between rapid convergence and the ability to generalize from features that are specifically relevant to the domain. The experimental outcomes not only validate the theoretical assumptions made at the inception of this study but also underscore the need for further fine-tuning strategies that optimize model performance without compromising the clinical integrity of the feature representations. These insights lay the groundwork for future research aimed at enhancing the robustness and applicability of transfer learning methodologies in medical imaging.

\section{Future Work}

The ultimate objective of this research is to develop clinically viable dermatological tools based on self-supervised learning models that improve diagnostic support in hospital settings. Once a robust model is achieved—matching or surpassing the performance of existing pretrained models such as ImageNet while minimizing overfitting—the next step involves deploying practical and efficient solutions through knowledge distillation. This process allows the creation of smaller (less parameters for overfitting), more efficient models optimized for specific clinical tasks, facilitating their integration into real-world workflows.

In practice, these distilled models could be tailored for various applications, such as:

\begin{itemize}
    \item \textbf{Melanoma Detection:} Specialized models for identifying melanoma at different stages and depths, aiding early diagnosis and reducing unnecessary biopsies.
    \item \textbf{BCC Pattern Recognition:} Tools capable of differentiating BCC subtypes (e.g., superficial, nodular, infiltrative), improving interpretability and decision-making for dermatologists detecting the different BCC patterns \cite{matas2024aidrivenskincancerdiagnosis}.
    \item \textbf{Actinic and Clinical Keratosis Detection:} Early identification of precancerous lesions, helping clinicians intervene before progression to squamous cell carcinoma.
    \item \textbf{General Dermatological Screening:} AI-driven pre-screening tools that prioritize high-risk cases and assist in triaging patients for further examination.
\end{itemize}

By fine-tuning these models for specific tasks, we aim to provide dermatologists with reliable, interpretable, and clinically relevant tools, ensuring that AI complements rather than replaces human expertise.

\subsection{Impact on Diagnostic Accuracy and Decision Support}

The successful implementation of these AI models in a clinical setting could significantly enhance diagnostic accuracy, particularly for conditions where subtle morphological differences define disease severity and treatment pathways. Given the high intra- and inter-observer variability in dermatological diagnosis, these models offer:

\begin{itemize}
    \item \textbf{Standardized, data-driven assessments}, reducing diagnostic discrepancies among clinicians. As demonstrated in \cite{silvaclavería2024concordancebasalcellcarcinoma}, diagnostic variability is highly dependent on the clinical specialty, highlighting the need for standardized AI-driven tools to ensure consistent and objective evaluations.
    \item \textbf{Augmented decision support}, allowing doctors to confirm or refine their evaluations based on AI-generated insights.
    \item \textbf{Faster and more scalable screening}, improving patient throughput in hospitals with limited dermatological specialists. Additionally, by enhancing the flow of information between primary care and specialists—such as with the proposed tool, which prioritizes lesions based on their clinical significance—the system optimizes patient management, ensuring that cases requiring urgent attention are promptly identified, thus improving efficiency in hospitals with limited dermatological expertise.

\end{itemize}

Additionally, by incorporating explainable AI (XAI) techniques, such as heatmaps or lesion attribution maps, these tools could improve transparency and trust among healthcare professionals, further strengthening their adoption in real-world medical practice.

\section{Acknowledgements}

This work was supported by the Andalusian Regional Government (PROYEXCEL\_00889) and is part of the project of the aid PID2021-127871OB-I00, funded by MCIN/AEI/10.13039/501100011033 and ERDF/European Union (NextGenerationEU/PRTR).

We would also like to express our gratitude to the medical staff of the ``Hospital Universitario Virgen Macarena'' for granting us access to the database used in this project. Their collaboration has been essential for the development of this study.

\bibliographystyle{IEEEtran}
\bibliography{references}

\end{document}